\documentclass{article}

 \usepackage[preprint]{neurips_2026}

\usepackage[utf8]{inputenc} 
\usepackage[T1]{fontenc}    
\usepackage{hyperref}       
\usepackage{url}            
\usepackage{booktabs}       
\usepackage{amsfonts}       
\usepackage{nicefrac}       
\usepackage{microtype}      
\usepackage{xcolor}         
\usepackage{caption}

\usepackage{tcolorbox}
\usepackage{multirow}
\usepackage{enumitem}
\usepackage{pifont}
\usepackage{makecell}
\usepackage{wrapfig}
\usepackage{amsmath}
\usepackage{amssymb}

\newcommand{\ie}[0]{{{\textit{i.e.}}}}
\newcommand{\eg}[0]{{{\textit{e.g.}}}}
\newcommand{\model}[0]{Efficient-VLN}

\newcommand{\myparagraph}[1]{\vspace{-2.mm} \paragraph{#1}}

\definecolor{navyblue}{HTML}{0071BC}
\definecolor{hotpink}{HTML}{FF0080}
\definecolor{oai-white}{HTML}{FFFFFF}
\definecolor{oai-black}{HTML}{000000}
\definecolor{oai-red}{HTML}{FF4500}
\definecolor{oai-green}{HTML}{51DA4C}
\definecolor{oai-blue}{HTML}{0000FF}
\definecolor{oai-yellow}{HTML}{FFF639}
\definecolor{oai-magenta}{HTML}{FF45FF}
\definecolor{oai-cyan}{HTML}{00FFFF}
\definecolor{oai-orange}{HTML}{FE7600}
\definecolor{oai-violet}{HTML}{8A2BE2}
\definecolor{oai-brown}{HTML}{A0522D}
\definecolor{oai-green-050}{HTML}{F4FFF4}
\definecolor{oai-green-100}{HTML}{E9FFE8}
\definecolor{oai-green-200}{HTML}{D9FFD8}
\definecolor{oai-green-300}{HTML}{C9FFC7}
\definecolor{oai-green-400}{HTML}{A6FFA3}
\definecolor{oai-green-500}{HTML}{7CF178}
\definecolor{oai-green-600}{HTML}{51DA4C}
\definecolor{oai-green-700}{HTML}{3FA93B}
\definecolor{oai-green-800}{HTML}{2D712A}
\definecolor{oai-green-900}{HTML}{193718}
\definecolor{oai-gray-000}{HTML}{FFFFFF}
\definecolor{oai-gray-100}{HTML}{FAFAFA}
\definecolor{oai-gray-200}{HTML}{F5F5F5}
\definecolor{oai-gray-300}{HTML}{E5E5E5}
\definecolor{oai-gray-400}{HTML}{FFB7A4}
\definecolor{oai-gray-500}{HTML}{CDCDCD}
\definecolor{oai-gray-600}{HTML}{A8A8A8}
\definecolor{oai-gray-700}{HTML}{747474}
\definecolor{oai-gray-800}{HTML}{393939}
\definecolor{oai-gray-900}{HTML}{000000}

\definecolor{mygreen}{HTML}{3cb44b}

\definecolor{myred}{HTML}{E33222}
\definecolor{gr}{RGB}{0, 146, 0}

\title{Efficient-VLN: A Simple yet Strong Baseline for Efficient Vision-Language Navigation}

\author{
Duo Zheng$^{1}$ \qquad
Shijia Huang$^{2}$ \qquad
Yanyang Li$^{2}$ \qquad
\textbf{Liwei Wang}$^{1}$ \\
$^1$The Chinese University of Hong Kong \quad 
$^2$Weitu AI 
}

\begin{document}

\maketitle

\begin{abstract}
While Multimodal Large Language Models (MLLMs) have demonstrated significant promise in Vision-Language Navigation (VLN), existing agents remain heavily constrained by systemic bottlenecks across inference, training, and data collection. Specifically, they suffer from prohibitive latency due to visual history reprocessing, action leakage during sequence-packed training, and suboptimal exploration in self-correction data collection. To overcome these intertwined challenges, we present \model{}, a highly efficient and robust baseline that systematically resolves these issues through three simple-yet-effective mechanisms. (1) Inference: We introduce \textit{KV-cache reuse with contiguous RoPE}, enabling the model to process only the newly observed frame at each step for real-time inference. (2) Training: We propose \textit{packed training with an action-isolating mask} to accelerate throughput while effectively bridging the training-inference gap by preventing action leakage. (3) Data Collection: We employ an \textit{Adaptive DAgger} to dynamically balance autonomous exploration and oracle guidance, enhancing error-recovery capability without escalating computational costs. 
Extensive evaluations show that \model{} significantly advances the state-of-the-art across the R2R-CE (73.2\% SR) and RxR-CE (75.6\% SR) benchmarks. Meanwhile, it yields a 28\% latency reduction compared to the previous state-of-the-art StreamVLN, establishing a new paradigm for streaming MLLM-based navigation.
\end{abstract}

\section{Introduction}
\label{sec:intro}
Building autonomous robots capable of navigating via language instructions is a fundamental pursuit in embodied AI. To this end, Vision-Language Navigation (VLN) \cite{anderson2018vision, krantz_vlnce_2020, ku2020room, qi2020reverie} has emerged as a pivotal research task, attracting extensive interest from the community \cite{cheng2024navila,zhang2025navfom,zhang2024uninavid,zhang2024navid,zheng2024navillm,zhou2024navgpt}. Driven by the unprecedented success of Multimodal Large Language Models (MLLMs) in generic multimodal understanding, there has been a growing trend toward adapting these powerful models for embodied navigation. By harnessing the extensive world knowledge inherent in MLLMs, this paradigm \cite{cheng2024navila, zhang2024navid,zheng2024navillm} has demonstrated superior capabilities and significant promise.

Despite this rapid progress, existing MLLM-based agents \cite{ long2024instructnav, wei2025streamvln, xue2025omninav, yu2025correctnav, zeng2025janusvln, zhang2024uninavid} remain heavily constrained by systemic bottlenecks across three fundamental dimensions: the inference pipeline, the training paradigm, and the data collection strategy. Specifically, prevailing approaches \cite{cheng2024navila, wei2025ground, wei2025streamvln} suffer from prohibitive latency that precludes real-time inference, suboptimal exploitation of supervision signals during training, and error-correction data that either offer marginal gains or rely excessively on external models. Below, we dissect the specific challenges within each dimension that the community has yet to fully resolve:

\noindent\textbf{1) Computational redundancy in inference.} A navigation agent observes only one new frame per step, yet most current methods reprocess the entire visual history, which leads to significant inference latency. For instance, NaVILA \cite{cheng2024navila} and CorrectNav \cite{yu2025correctnav} require reprocessing 8 and 16 frames respectively, incurring a massive token footprint at each decision step. 
Recent work, such as StreamVLN \cite{wei2025streamvln}, alleviates this with a global memory and a sliding window but still requires recomputing the global memory whenever the sliding window reaches the boundary, thereby triggering periodic latency spikes that hinder smooth deployment.

\noindent\textbf{2) Optimization bottlenecks in the training paradigm.} Conventional training regimes typically supervise the agent to predict only a short horizon of future actions at each step, underutilizing the rich temporal dynamics across the full visual history. To mitigate this, a common workaround involves packing multiple decision steps from a single trajectory into one unified training sequence, akin to the multi-turn dialog format used by StreamVLN \cite{wei2025streamvln}. However, this introduces an inherent limitation during training: once ground-truth actions from earlier steps become visible to later steps, the model suffers from action leakage. Consequently, the agent is prone to exploiting spurious correlations between past and future actions \cite{cen2025worldvla}. This fundamental training-inference mismatch inevitably exacerbates compounding errors during actual inference, where prior ground-truth actions are distinctly unavailable.

\noindent\textbf{3) Inefficient exploration in data collection.} Standard imitation learning relies heavily on expert demonstrations, inevitably suffering from exposure bias due to distributional shifts during inference. To mitigate this, recent approaches \cite{wei2025streamvln, zhang2024navid} collect self-correction data via the DAgger \cite{ross2011reduction} algorithm by mixing the learned policy with oracle guidance. However, such rigid mixing schedules inherently impose a suboptimal trade-off between exploration diversity and sample efficiency, an underexplored bottleneck in VLN. Furthermore, while methods like CorrectNav \cite{yu2025correctnav} introduce heuristic pipelines to identify failure points using off-the-shelf MLLMs, relying on external models also carries the risk of propagating cascading errors.

To address these limitations, we introduce \model{}, a simple yet strong VLN baseline that systematically tackles the aforementioned bottlenecks in inference efficiency, training optimization, and data collection.
\noindent\textbf{(i) Inference: KV-cache reuse with contiguous RoPE.} To bypass the redundancy of repetitive visual encoding, we cache the key-value (KV) tensors before applying Rotary Position Embedding (RoPE) and selectively retain useful historical states via a frame sampler. These cached tensors are dynamically concatenated with the newly encoded frame, followed by a rotation transformation using contiguous RoPE IDs. This allows the model to process only a single new frame at each step, yielding near-constant per-step latency for real-time streaming.
\noindent\textbf{(ii) Training: Packed training with an action-isolating mask.} We pack multiple decision steps from one trajectory into a single training sample sharing a common visual prefix. This enables accelerating the training throughput without arbitrarily escalating the sequence length. To prevent action leakage, we employ a customized action-isolating mask that hides the supervised actions of all preceding steps from each prediction. Consequently, each supervised step receives exactly the same context it would encounter during inference.
\noindent\textbf{(iii) Data Collection: Adaptive DAgger.} We replace the conventional time-invariant mixing schedule with a time-decaying schedule. Specifically, it encourages the agent to explore autonomously during the early stages of a rollout, where self-correction data is most valuable, and gradually defers to oracle guidance in later steps. This design efficiently concentrates exploration in crucial steps while strictly bounding the length of trajectories.

We evaluate our \model{} on two widely used benchmarks, \ie, R2R-CE \cite{krantz_vlnce_2020} and RxR-CE \cite{ku2020room}. 
\model{} sets new state-of-the-art performance, achieving a \textbf{73.2\%} Success Rate (SR) on R2R-CE and \textbf{75.6\%} SR on RxR-CE. Notably, it outperforms the strongest baseline OmniNav \cite{xue2025omninav}, which relies on panoramic observations, by absolute margins of \textbf{+3.7\%} and \textbf{+2.0\%}, respectively. 
At the same time, it operates at \textbf{159ms/chunk}, illustrating an \textbf{82\%} reduction over the same backbone without cache (898ms) and \textbf{28\%} faster than StreamVLN (221ms) on identical hardware.

In summary, our main contributions are threefold:
\begin{itemize}[itemsep=2pt, topsep=2pt, parsep=0pt, leftmargin=*]
\item We present \model{}, a highly efficient and robust baseline for MLLM-based Vision-Language Navigation that comprehensively resolves systemic bottlenecks across inference latency, training optimization, and data collection.
\item We introduce three simple-yet-effective mechanisms highly tailored for streaming VLN: (1) \textit{KV-cache reuse with contiguous RoPE} to eliminate computational redundancy; (2) \textit{packed training with an action-isolating mask} to accelerate throughput while preventing action leakage; and (3) an \textit{Adaptive DAgger} for correction data collection which enhances error-recovery capability while strictly bounding trajectory length.
\item Extensive evaluations demonstrate that \model{} establishes new state-of-the-art results on both R2R-CE (73.2\% SR) and RxR-CE (75.6\% SR), even outperforming the strongest panoramic-view baselines. Furthermore, it operates at an ultra-low latency of 159\,ms/step, achieving a 28\% speedup over the previous state-of-the-art StreamVLN.
\end{itemize}

\section{Related Work}

\myparagraph{Vision-Language Navigation (VLN).}
VLN~\cite{anderson2018vision, ku2020room, qi2020reverie, thomason2020vision} stands as a fundamental task for Embodied AI, drawing extensive attention from the research community.
Early efforts in VLN primarily explored various techniques, such as specialized memory structures~\cite{chen2022think, hong2022bridging, wang2023gridmm}, data augmentation~\cite{tan2019envdrop, wang2024sim, wang2023scalevln}, and reinforcement learning~\cite{jain2019stay, wang2019reinforced}.
More recently, driven by the remarkable success of Multimodal Large Language Models (MLLMs), there has been a paradigm shift toward leveraging these powerful foundation models to address this task~\cite{cheng2024navila, wang2025dynam3d,wei2025ground, wei2025streamvln, xue2025omninav, yu2025correctnav, zhang2024navid, zheng2024navillm}.
NaviLLM~\cite{zheng2024navillm} and NaVid~\cite{zhang2024navid} pioneered the development of trainable MLLM-based models in embodied navigation.
Subsequently, Uni-NaVid~\cite{zhang2024uninavid} extended the research scope to a broader range of embodied AI tasks, while NaVILA~\cite{cheng2024navila} introduced hierarchical architectures that decouple high-level planning from low-level execution.
Despite these advancements, the training of MLLM-based agents remains bottlenecked by two critical issues, \emph{i.e.}, training inefficiencies and data effectiveness.
In this paper, we systematically address these challenges by proposing a packed training paradigm with an action-isolating mask to prevent action leakage, coupled with an Adaptive DAgger to collect highly effective self-correction data.

\myparagraph{Efficient Vision-Language Navigation.}
Early efforts toward efficient VLN primarily focused on optimizing smaller, specialized architectures through techniques such as knowledge distillation~\cite{2025MiniVLN}, specialized memory structures~\cite{hong2021vln}, and heuristic exploration on topological maps~\cite{an2023etpnav, chen2022think}.
More recently, while MLLM-based agents have achieved superior performance and generalizability, they suffer from significant inference latency due to the redundant processing of extensive visual history at each time step.
To mitigate this, several acceleration techniques have been explored.
For instance, NaVid~\cite{zhang2024navid} and Uni-NaVid~\cite{zhang2024uninavid} reduce token usage through compression, whereas NaVILA~\cite{cheng2024navila} employs parameter quantization, at the expense of navigation performance.
The work most closely related to ours is StreamVLN~\cite{wei2025streamvln}, which reuses a previously encoded KV cache within a sliding window but must recompute the global memory once the window boundary is reached
Consequently, such approaches are constrained by periodic computational spikes when refreshing the global memory.
In stark contrast, by reusing unrotated KV tensors with contiguous RoPE, our architecture eliminates these spikes, leading to near-constant per-step inference latency.

\myparagraph{Streaming Video Understanding.}
Streaming video understanding~\cite{chen2024videollm, chen2026streamingtom, di2025streaming, xu2025streamingvlm, yao2025timechat} is a pivotal capability for embodied AI agents, requiring the model to process continuous visual streams with minimal latency.
Existing approaches primarily concentrate on alleviating computational burdens by reducing token redundancy through compression and merging techniques~\cite{chen2026streamingtom, yao2025timechat}.
Concurrently, another line of work investigates the efficiency gains from reusing previously computed KV caches to avoid redundant reprocessing~\cite{di2025streaming, xu2025streamingvlm}.
While these methods are predominantly optimized for real-time Video-QA benchmarks (\eg, sports broadcasts and live streams), they often lack the mechanisms required for complex navigation. 
In contrast, our approach is specifically tailored to navigation tasks by decoupling training-inference sampling strategies and introducing efficient memory management.

\begin{figure*}[t] 
\centering 
\includegraphics[width=1.0\textwidth]{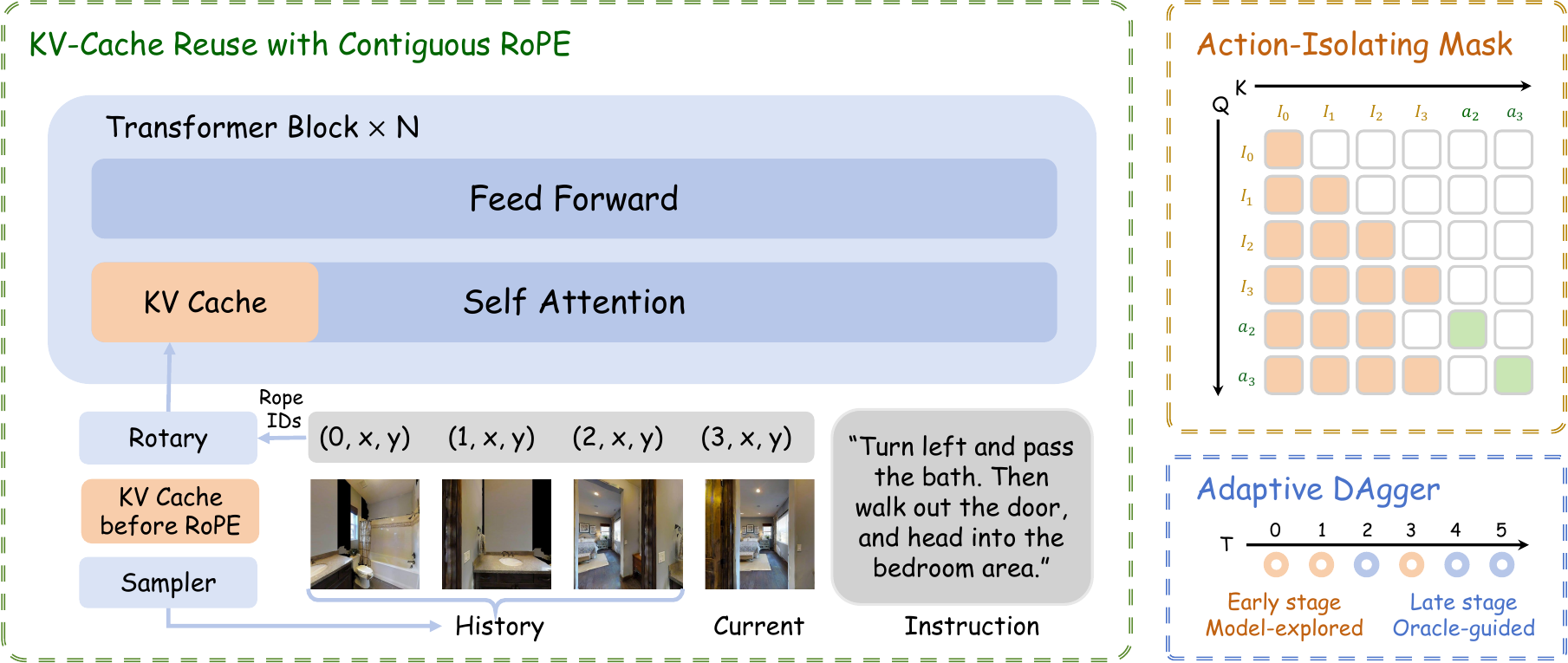}
\caption{\textbf{Overview of \model{} framework.}
\textbf{(Left) KV-Cache Reuse with Contiguous RoPE:} 
KV tensors before rotation are selected and concatenated with the newly encoded frame, and subsequently rotated using contiguous RoPE IDs.
\textbf{(Top right) Action-Isolating Mask}:
Visual tokens attend exclusively to preceding visual context (orange blocks). To prevent leakage, each action token attends only to its corresponding visual history and its current action chunk (green blocks), strictly masking out ground-truth actions from prior steps.
\textbf{(Bottom right) Adaptive DAgger}:
Within each rollout, the agent initially explores using its own policy (orange circles) and gradually defers to oracle guidance in later steps (blue circles).
}

\label{fig:method}
\end{figure*}

\section{Method}

\subsection{Task Formulation}
In the Vision-Language Navigation (VLN) task, a model is tasked with navigating to a target location following a language instruction $I$. 
At each step $t$, the model receives a new image $v_t$ as the current observation and predicts actions conditioned on the instruction $I$ and the full observation history $\{v_1, v_2, \cdots, v_t\}$.
Following previous works \cite{zhang2024uninavid, wei2025streamvln}, the model is asked to predict an action chunk consisting of a sequence of four future actions. Each individual action within the chunk is selected from a discrete action space comprising \texttt{forward}, \texttt{left}, \texttt{right}, and \texttt{stop}.
An episode terminates when the model produces the \texttt{stop} action.

Our model is built upon the Qwen3-VL-8B architecture \cite{bai2025qwen3}. 
Figure \ref{fig:method} illustrates the three core mechanisms of \model{}. 
Specifically, we first elevate inference efficiency by reusing the KV cache with contiguous RoPE (\S\ref{sec:method_kv}). 
We then devise a packed training paradigm equipped with an action-isolating mask to eliminate action leakage (\S\ref{sec:method_pack}). 
Finally, we introduce an Adaptive DAgger strategy to structurally enhance self-correction data collection (\S\ref{sec:method_dagger}).

\subsection{KV Cache Reuse with Contiguous RoPE}
\label{sec:method_kv}

At step $t$, an MLLM-based VLN agent conditions its predictions on the instruction $I$ and a subset $\mathcal{S}_t \subseteq \{v_1,\dots,v_t\}$ selected by a frame sampler.
While KV cache reuse is a standard practice for inference acceleration, it introduces a critical bottleneck in navigation scenarios.
Specifically, cached key tensors are heavily entangled with their absolute positions through RoPE.
Consequently, a mismatch occurs between the pre-computed cache and the token's new position whenever the new state updates the context window (\eg, via re-sampling or re-ordering).
Due to this constraint, prior approaches for streaming VLN like StreamVLN \cite{wei2025streamvln} resort to periodically recomputing the global memory once the sliding window reaches its boundary.

To address this, we draw inspiration from StreamingVLM \cite{xu2025streamingvlm} and store the KV cache prior to the rotation transformation.
At each step, we identify relevant previously seen frames using a specified sampler and dynamically retrieve their unrotated KV tensors.
These selected KV tensors before RoPE are then concatenated with the newly encoded frame.
These historical tensors are then concatenated with those of the newly encoded frame. Importantly, we assign contiguous position IDs to this combined sequence and apply the RoPE transformation on the fly. 
This design allows the agent to process only the current frame for the vision encoder and transformer prefilling, yielding near-constant per-step latency.
For the frame sampler, we investigate two strategies for $\mathcal{S}_t$\footnote{Empirically, uniform sampling achieves superior performance, which we attribute to its broad coverage of the history. Consequently, we employ this as our default configuration in all subsequent experiments unless otherwise specified.
}, \ie, \emph{uniform sampling} which uniformly samples a fixed number of frames across the entire history, and \emph{sliding window-based sampling} which retains the most recent frames using a stride of $4$.

\myparagraph{Decoupled Frame Sampling.}
Prior methods \cite{zhang2024uninavid, wei2025streamvln} restrict inference-time sampling to match the training configuration.
Our cache mechanism decouples position IDs from the KV cache. This enables flexible adaptation to arbitrary sampling strategies during inference. 
Empirically, we observe that applying different sampling strategies during inference with the same trained model yields marginal performance variance (\S \ref{sec:exp_component}).
This demonstrates the robustness of our method, allowing adaptive adoption of the sampling strategy according to task requirements.

\myparagraph{Memory Management.}
While the caching mechanism resolves computational bottlenecks, retaining a growing history inevitably introduces GPU memory constraints.
To this end, we offload the cache to the CPU and reload only the segments selected by the sampler, which keeps GPU memory bounded as the path grows.
For a context of $N$ historical frames, the sliding-window strategy needs to transfer one new segment per step, while uniform sampling transfers $N/2$ on average. Both are negligible compared to the cost of re-encoding all $N$ frames from scratch.

We acknowledge that reusing these cached tensors is mathematically an approximation. In a causal LLM, the KV tensors are intrinsically entangled with the specific historical context present during their initial prefilling, which may differ from the newly sampled context $\mathcal{S}_t$. Nevertheless, we justify this approximation empirically, as it yields substantial efficiency gains without compromising navigation performance (\S \ref{sec:exp_component}).

\subsection{Packed Training via Action Isolation}
\label{sec:method_pack}

Existing methods \cite{cheng2024navila, wei2025streamvln} consume thousands of tokens for each action chunk prediction during training, yielding prohibitive computational inefficiencies.
To mitigate this, a practical solution employed by StreamVLN \cite{wei2025streamvln} involves packing multiple observation-action pairs from the same trajectory into an aggregated training sequence.
However, there is a risk that such implementation will cause error propagation as earlier incorrect predictions inevitably influence future actions \cite{cen2025worldvla}.

Consider a packed sequence containing $N$ frames $\{v_1,\dots,v_N\}$ together with the corresponding ground truth action chunks $\{a_1,\dots,a_N\}$.
We supervise only the last $M \le N$ action chunks $\{a_{N-M+1},\dots,a_{N}\}$. 
The earlier $N-M$ frame is responsible for providing the history context.
To prevent action leakage, we replace the standard causal mask with an action-isolating mask (Fig. \ref{fig:method}).
Under this mask, visual tokens are restricted to attending only to the instruction and preceding visual frames. Concurrently, when predicting the action chunk $a_t$ at step $t$, the tokens attend exclusively to the instruction, the visual history up to $v_t$, and the preceding tokens within the current chunk itself.

Conceptually, a packed sequence is akin to $M$ parallel single-step training examples sharing the same sequence prefix. 
This design effectively enlarges the batch size without escalating token consumption, thereby boosting training throughput and stabilizing optimization. 
Furthermore, the action-isolating mask strictly aligns the training objective with the inference setup, where historical actions are unavailable. By strictly masking out prior actions, it completely eliminates the risk of error propagation from earlier incorrect predictions.

\subsection{Adaptive DAgger}
\label{sec:method_dagger}

DAgger \cite{ross2011reduction} mitigates exposure bias by rolling out a mixture of the learned policy $\hat{\pi}$ and an oracle policy $\pi^*$, then relabeling visited states with oracle actions.
Conventional implementations use a fixed mixing probability $\beta$. 
At every step of every rollout, $\pi^*$ is followed with probability $\beta$ and $\hat{\pi}$ otherwise.
This treats all steps as equally important, which ignores the sequential nature of navigation.
Errors made early in a trajectory have cascading effects and significantly affect the subsequent state.
Conversely, mistakes made near the end have a solely limited influence.

\myparagraph{Time-Decaying Schedule for Policy Exploration.}
We therefore replace the constant $\beta$ with a time-decaying schedule that decreases the policy's exploration probability monotonically over the rollout. 
This is formulated as
\begin{equation}
P_{\text{model}} = 1 - \beta_t = \alpha^{t/T},
\end{equation}
where $t$ is the current step, $T$ is the length of the ground-truth path, and $\alpha \in (0,1)$ is a decay weight.
At $t=0$ we have $\beta_t=0$, so the rollout starts purely on the learned policy and exposes the agent to explored states.
As $t$ increases, the term $\beta_t$ gradually approaches 1. 
This allows the oracle to pull the agent toward the goal and cap the rollout length.
This simple modification yields a measurable effect (Sec.~\ref{sec:exp_dagger}). 
Compared to the best constant-$\beta$ baseline, the adaptive schedule achieves higher SR and SPL while collecting noticeably shorter rollouts, which in turn shortens inference steps.

\section{Experiments}
\label{sec:experiments}

\subsection{Experimental Setup}
\label{sec:setup}

\myparagraph{Benchmarks.}
We evaluate our method on two widely used VLN benchmarks, \ie, R2R-CE \cite{krantz_vlnce_2020} and RxR-CE \cite{ku2020room}, both based on the Habitat simulator.
R2R-CE comprises 10,819 episodes for training and 1,839 for the val-unseen split.
RxR-CE includes 19,996 episodes for training and 3,669 for the val-unseen split.

\myparagraph{Metrics.}
We employ the following metrics:
(1) Navigation Error (\textbf{NE}), which measures the average distance between the agent's final location and the destination;
(2) Success Rate (\textbf{SR}), the percentage of the episodes where the agent stops within a predefined distance threshold of the target;
(3) Oracle Success Rate (\textbf{OS}), the SR given an oracle stopping policy;
(4) Success rate weighted by the Path Length (\textbf{SPL}), the SR weighted by the ratio of the ground truth length to the agent's path length.
We additionally report \textbf{per-step latency} (ms) to measure the inference efficiency.

\myparagraph{Environment Setup.}
Both the data collection and evaluation processes are conducted within the Habitat simulator \cite{savva2019habitat}.
For the navigation task, we adopt the configuration from NaVILA \cite{cheng2024navila}, setting the visual observation resolution to $512 \times 512$ and the horizontal field of view (HFOV) to $90^\circ$. 
Each navigation episode terminates either when the agent explicitly outputs a \texttt{STOP} action or when its step count reaches the maximum limit of 500.
All latency numbers are measured on the same single H200 GPU with batch size 1, FP16 inference, and an identical software stack across all compared methods.

\myparagraph{Implementation Details.}
We build \model{} upon the Qwen3-VL-8B architecture \cite{bai2025qwen3}.
Our training pipeline consists of two distinct phases, \ie, policy initialization for self-correction data collection, and final model training. 
First, to enable our Adaptive Rollout Strategy, we train an initial policy on the combined R2R-CE \cite{krantz_vlnce_2020} and RxR-CE \cite{ku2020room} datasets, which consist of 30K trajectories in total. 
We then deploy this initial model to collect another 30K self-correction trajectories. 
While this preliminary model attains modest performance, we observe significant improvements after incorporating its early-exploration-rich data into the next phase. 
Subsequently, in the final training phase, we fully fine-tune \model{} on a comprehensive mixed dataset comprising 106K samples: the initial R2R-CE and RxR-CE data (30K), the collected rollout trajectories (30K), and a subset of ScaleVLN \cite{wang2023scalevln} (46K). 

\begin{table*}[t]
\centering
\small
\resizebox{1.\linewidth}{!}{
\setlength{\tabcolsep}{2.mm}  
\begin{tabular}{l|cccc|cccc|ccc} 
\toprule
\multirow{2}{*}{Method} & \multicolumn{4}{c|}{Observation Encoder} & \multicolumn{4}{c|}{R2R-CE Val-Unseen} & \multicolumn{3}{c}{RxR-CE Val-Unseen} \\ 
\cmidrule(lr){2-5} \cmidrule(lr){6-9} \cmidrule(lr){10-12} 
      & Pano. & Odo. & Depth & S.RGB & NE$\downarrow$ & OS$\uparrow$ & SR$\uparrow$ & SPL$\uparrow$ & NE$\downarrow$ & SR$\uparrow$ & SPL$\uparrow$ \\ 
\midrule

HPN+DN$^*$~\cite{krantz2021waypoint} & $\checkmark$ & $\checkmark$ & $\checkmark$ &  & 6.31  & 40.0  & 36.0  & 34.0  & - & - & - \\

CMA$^*$~\cite{hong2022bridging}      & $\checkmark$ & $\checkmark$ & $\checkmark$ &  & 6.20  & 52.0  & 41.0  & 36.0  & 8.76 & 26.5 & 22.1 \\

VLN$\protect\circlearrowright$BERT$^*$~\cite{hong2022bridging} & $\checkmark$ & $\checkmark$ & $\checkmark$ &  & 5.74  & 53.0  & 44.0  & 39.0  & 8.98 & 27.0 & 22.6 \\

Sim2Sim$^*$~\cite{krantz2022sim}    & $\checkmark$ & $\checkmark$ & $\checkmark$ &  & 6.07  & 52.0  & 43.0  & 36.0  & - & - & - \\

GridMM$^*$~\cite{wang2023gridmm}    & $\checkmark$ & $\checkmark$ & $\checkmark$ &  & 5.11  & 61.0  & 49.0  & 41.0  & - & - & - \\

ETPNav$^*$~\cite{an2023etpnav}      & $\checkmark$ & $\checkmark$ & $\checkmark$ &  & 4.71  & 65.0  & 57.0  & 49.0  & 5.64 & 54.7 & 44.8 \\ 

ScaleVLN$^{*}$~\cite{wang2023scalevln} & $\checkmark$ & $\checkmark$ & $\checkmark$ &  & 4.80  & --    & 55.0  & 51.0  & -& - & - \\
\midrule

InstructNav~\cite{long2024instructnav} & \checkmark & \checkmark & \checkmark &  & 6.89 & --   & 31.0 & 24.0 & - & - & - \\

Seq2Seq~\cite{krantz_vlnce_2020} & $\checkmark$  & & $\checkmark$ & & 7.77 & 37.0  & 25.0  & 22.0  & 12.10 & 13.9 & 11.9 \\

CMA~\cite{krantz_vlnce_2020} & $\checkmark$  & & $\checkmark$ & & 7.37 & 40.0  & 32.0  & 30.0  & -- & -- & -- \\

AG-CMTP \cite{chen2021topological} & $\checkmark$ & $\checkmark$ & $\checkmark$ &  & 7.90  & 39.0  & 23.0  & 19.0  & - & - & - \\

R2R-CMTP~\cite{chen2021topological}  & $\checkmark$ & $\checkmark$ & $\checkmark$ &  & 7.90  & 38.0  & 26.4  & 22.7  & - & - & - \\

LAW~\cite{raychaudhuri2021language}       &  & $\checkmark$ & $\checkmark$ & $\checkmark$ & 6.83  & 44.0  & 35.0  & 31.0  & 10.90 & 8.0 & 8.0 \\

CM2~\cite{georgakis2022cross}        &  & $\checkmark$ & $\checkmark$ & $\checkmark$ & 7.02  & 41.5  & 34.3  & 27.6  & - & - & - \\

WS-MGMap~\cite{chen2022weakly}       &  & $\checkmark$ & $\checkmark$ & $\checkmark$ & 6.28  & 47.6  & 38.9  & 34.3  & - & - & - \\

ETPNav + FF~\cite{wang2024sim}  &  & $\checkmark$ & $\checkmark$ & $\checkmark$ & 5.95  & 55.8  & 44.9  & 30.4  & 8.79 & 25.5 & 18.1 \\

Dynam3D \cite{wang2025dynam3d} &  & $\checkmark$ & $\checkmark$ & $\checkmark$ & 5.34 & 62.1 & 52.9 & 45.7 & - & - & - \\

D3D-VLP \cite{wang2025d3d} &  & $\checkmark$ & $\checkmark$ & $\checkmark$ & 4.73 & 67.2 & 61.3 & 56.1 \\

NavFoM (Four Views)~\cite{zhang2025navfom} & $\checkmark$ & & & & 5.01 & 64.9 & 56.2 & 51.2 & 5.51 & 57.4 & 49.4 \\

OmniNav~\cite{xue2025omninav} & $\checkmark$ & & &  & \underline{3.74} & \underline{74.6} & \underline{69.5} & \textbf{66.1} & \underline{3.77} & 73.6 & \underline{62.0} \\

\midrule

NaVid~\cite{zhang2024navid}   &  &  &  & $\checkmark$ & 5.47  & 49.1  & 37.4  & 35.9  & - & - & - \\

NaVILA~\cite{cheng2024navila}   &  &  &  & $\checkmark$ & 5.22  & 62.5  & 54.0  & 49.0  & 6.77 & 49.3 & 44.0 \\

UniNaVid~\cite{zhang2024uninavid}   &  &  &  & $\checkmark$ & 5.58  & 53.3  & 47.0  & 42.7  & 6.24 & 48.7 & 40.9 \\
DecoVLN~\cite{xin2026decovln} &  &  &  & $\checkmark$ & 5.01 & 63.5 & 56.3 & 50.5 & 5.73 & 54.2 & 46.3 \\
StreamVLN~\cite{wei2025streamvln}   &  &  &  & $\checkmark$ & 4.98  & 64.2  & 56.9  & 51.9  & 6.22 & 52.9 & 46.0 \\
NavFoM (Single View) \cite{zhang2025navfom} & & & &  $\checkmark$ & 5.01 & 64.9 & 56.2 & 51.2 & 5.51 & 57.4 & 49.4 \\
JanusVLN \cite{zeng2025janusvln} & & & & $\checkmark$ & 4.78 & 65.2 & 60.5 & 56.8 & 6.06 & 56.2 & 47.5 \\
DualVLN \cite{wei2025ground} & & & & $\checkmark$ & 4.05 & 70.7 & 64.3 & 58.5  & 4.58 &  61.4 & 51.8 \\
CorrectNav \cite{yu2025correctnav} & & & & $\checkmark$ & 4.24 & 67.5 & 65.1 & 62.3 & 4.09 & 69.3 & \textbf{63.3} \\

\textbf{\model{}} & & & & \checkmark & \textbf{3.29} & \textbf{78.2} &\textbf{73.2} & \underline{65.0} & \textbf{3.18} &\textbf{75.6 }& 60.0  \\
\bottomrule
\end{tabular}
}
\caption{
\textbf{Comparison with state-of-the-art methods on R2R-CE and RxR-CE Val-Unseen split.}
The ``Observation Encoder'' inputs include panoramic (Pano.), odometry (Odo.), depth image (Depth), and single RGB image (S.RGB).
$*$ denotes methods utilizing the waypoint predictor from~\cite{hong2022bridging}.
}
\label{tab:main_result}
\end{table*}

We employ the Adam optimizer with an effective batch size of 64 and a warmup ratio of 0.03. 
The learning rate is set to peak at $2 \times 10^{-5}$ after the warmup period, followed by a linear decay to zero. 
During training, the visual encoder remains frozen, while the MLLM backbone is fully updated. 
We keep $N=16$ historical frames in context, supervise $M=4$ actions per sample, and set the Adaptive Rollout decay weight $\alpha=0.6$. 
The entire training process consumes 544 GPU-hours (\ie, 34 hours of wall-clock time on a 2$\times$8 H200 GPU cluster).

\subsection{Comparison with State-of-the-Art}
\label{sec:exp_sota}

\noindent\textbf{\model{} sets a new SOTA on both benchmarks.}
Table~\ref{tab:main_result} presents a comparison with state-of-the-art methods on the R2R-CE and RxR-CE Val-Unseen benchmarks.
These methods can be broadly categorized into three groups:
(1) VLN models that employ an off-the-shelf waypoint predictor for candidate generation;
(2) models that utilize panoramic views or a combination of odometry, depth, and egocentric images; and
(3) monocular-based models that rely solely on single-view RGB input.

Under the monocular setting, our method achieves 3.29 NE and 73.2\% SR on R2R-CE, and 3.18 NE and 75.6\% SR on RxR-CE, improving over the previous best monocular method CorrectNav \cite{yu2025correctnav} by \textbf{8.1} SR points on R2R-CE and \textbf{6.3} on RxR-CE.
Although CorrectNav reports a marginally higher SPL on RxR-CE, this is because \model{} successfully navigates more complex episodes (driving our $+6.3$ SR gain). These challenging episodes require more exploration steps, which inherently lowers the overall SPL.
Furthermore, despite consuming only single-view RGB, \model{} outperforms the strongest panoramic baseline OmniNav \cite{xue2025omninav} by 3.7 SR on R2R-CE and 2.0 SR on RxR-CE.

\begin{table*}[t]
\centering
\setlength{\tabcolsep}{2mm} 
\resizebox{1.0\linewidth}{!}{
\begin{tabular}{@{} clc|c|cccc|ccc @{}} 
\toprule
\multirow{2}{*}{\#} & \multirow{2}{*}{Method} & \multirow{2}{*}{Use Cache} & \multirow{2}{*}{Latency$\downarrow$} & \multicolumn{4}{c}{R2R-CE Val-Unseen} & \multicolumn{3}{c}{RxR-CE Val-Unseen} \\
& & & & NE$\downarrow$ & OS$\uparrow$ & SR$\uparrow$ & SPL$\uparrow$ & NE$\downarrow$ & SR$\uparrow$ & SPL$\uparrow$ \\ 
\midrule
1 & StreamVLN \cite{wei2025streamvln} & \ding{51} & 221 & 4.98 & 64.2 & 56.9 & 51.9 & 6.22 & 52.9 & 46.0 \\ 
2 & DualVLN \cite{wei2025ground} & \ding{55} & 472 & 4.05 & 70.7 & 64.3 & 58.5  & 4.58 &  61.4 & 51.8 \\ 
\midrule
3 & \model{} & \ding{51} & \textbf{159} & \textbf{3.29} & \textbf{78.2} &\textbf{73.2} & \textbf{65.0} & \textbf{3.18} &\textbf{ 75.6 }& \textbf{60.0}  \\
\bottomrule
\end{tabular}
}
\caption{\textbf{Comparison of inference latency and navigation performance with state-of-the-art methods.}
Our model is evaluated using a uniform sampling strategy with 16 frames.
All these latency numbers are evaluated on 200 samples from R2R-CE using the same H200 GPU. 
For StreamVLN and DualVLN, we utilize their official codebase and model weights to measure the latency numbers.
}
\label{tab:latency}
\end{table*}

\begin{table*}[t]
\centering

\setlength{\tabcolsep}{2.mm}
\resizebox{1.0\linewidth}{!}{
\begin{tabular}{@{} c|ccc|c|cccc|ccc @{}}
\toprule
\multirow{2}{*}{\#}
 & \multicolumn{3}{c|}{Components}
 & \multirow{3}{*}{\makecell{Latency\\(ms)$\downarrow$}}
 & \multicolumn{4}{c|}{R2R-CE Val-Unseen}
 & \multicolumn{3}{c}{RxR-CE Val-Unseen} \\
 \cmidrule(lr){6-9} \cmidrule(lr){10-12} 
 & \makecell{Ada.\\DAgger} & \makecell{Packed\\Training} & \makecell{KV Cache\\Reuse}
 &
& NE$\downarrow$ & OS$\uparrow$ & SR$\uparrow$ & SPL$\uparrow$
 & NE$\downarrow$ & SR$\uparrow$ & SPL$\uparrow$ \\
\midrule
1 & \ding{55} & \ding{55} & \ding{51} & 159 & 4.12 & 69.8 & 60.3 & 51.5 & 3.96 & 63.8 & 51.1 \\
2 & \ding{51} & \ding{55} & \ding{51} & 159 & 3.57 & 75.7 & 67.8 & 56.7 & 3.11 & \textbf{75.6} & \textbf{61.7} \\
3 & \ding{51} & \ding{51} & \ding{55} & 898 & \textbf{3.22} & \textbf{78.3} & 72.5 & 64.1 & \textbf{3.03} & \textbf{75.6} & 61.5 \\
4 & \ding{51} & \ding{51} & \ding{51} & \textbf{159} & 3.29 & 78.2 & \textbf{73.2} & \textbf{65.0} & 3.18 & \textbf{75.6} & 60.0 \\
\bottomrule
\end{tabular}
}
\caption{
\textbf{Ablation of Efficient-VLN components.}
All variants are trained on the same combined dataset of R2R, RxR, DAgger, and ScalVLN for 30K steps.
The Adaptive DAgger and action-isolated packed training significantly improve navigation performance, while KV cache reuse with contiguous RoPE provides an 82\% latency reduction without compromising accuracy.
}
\label{tab:component_ablation}
\end{table*}

\begin{table*}[t]
\centering
\setlength{\tabcolsep}{2.5mm} 
\resizebox{1.0\linewidth}{!}{
\begin{tabular}{@{} c|cc|cccc|ccc @{}} 
\toprule
\multirow{2}{*}{\#} & \multirow{2}{*}{Train} & \multirow{2}{*}{Test} & \multicolumn{4}{c}{R2R-CE Val-Unseen} & \multicolumn{3}{c}{RxR-CE Val-Unseen} \\
 \cmidrule(lr){4-7} \cmidrule(lr){8-10}
& & & NE$\downarrow$ & OS$\uparrow$ & SR$\uparrow$ & SPL$\uparrow$ & NE$\downarrow$ & SR$\uparrow$ & SPL$\uparrow$ \\ 
\midrule
1 & \multirow{2}{*}{Sliding Window} & Sliding Window & 3.89 & 71.8 & 66.6 & 57.6 & 3.94 & 71.3 & 56.8 \\
2 & & Uniform & 4.07 & 70.4 & 63.8 & 55.6 & 4.16 & 68.2 & 54.3 \\ \midrule
3 & \multirow{2}{*}{Uniform} & Sliding Window & 3.34 & 77.5 &  72.2  & 64.1 & 3.24 & 74.1 & 59.3 \\
4 & & Uniform & \textbf{3.29} & \textbf{78.2} & \textbf{73.2} & \textbf{65.0} & \textbf{3.18} & \textbf{75.6} & \textbf{60.0} \\
\bottomrule
\end{tabular}
}
\caption{
\textbf{Effect of frame-sampling strategy on \model{}.} 
We train two variants using uniform and sliding-window sampling, respectively, and cross-evaluate each variant under both strategies.
}
\label{tab:sampling}
\end{table*}


\begin{table*}[t]
\centering
\setlength{\tabcolsep}{2.5mm} 
\resizebox{1.0\linewidth}{!}{
\begin{tabular}{@{} c|l|cccc|ccc @{}} 
\toprule
\multirow{2}{*}{\#} & \multirow{2}{*}{Training Scheme} & \multicolumn{4}{c|}{R2R-CE Val-Unseen} & \multicolumn{3}{c}{RxR-CE Val-Unseen} \\
\cmidrule(lr){3-6} \cmidrule(lr){7-9} 
& & NE$\downarrow$ & OS$\uparrow$ & SR$\uparrow$ & SPL$\uparrow$ & NE$\downarrow$ & SR$\uparrow$ & SPL$\uparrow$ \\
\midrule
1 & Single-step & 3.57 & 75.7 & 67.8 & 56.7 & 3.11 & \textbf{75.6 }&\textbf{ 61.7} \\
2 & Packed + causal mask & 4.31 & 75.8 & 62.3 & 50.8 & 4.31 & 61.8 & 48.1 \\
3 & Packed + action-isolating mask (ours) & \textbf{3.29} & \textbf{78.2} & \textbf{73.2} & \textbf{65.0} & \textbf{3.18} & \textbf{75.6} & 60.0 \\
\bottomrule
\end{tabular}
}
\caption{
\textbf{Effect of Packed Training via Action Isolation.}
``Single-step'' employs single-step supervision with a causal mask.
``Packed + causal mask'' adopts naive packing to format the data as a multi-turn dialog.
``Packed + action-isolating mask'' utilizes our proposed packing strategy and the action-isolation mask to eliminate action leakage. All three variants are trained for exactly 30K iterations to evaluate their performance.
}
\label{tab:attention_mask}
\end{table*}

\subsection{Analysis}

\myparagraph{Inference Latency.}
We evaluate inference efficiency by averaging the per-step latency over 200 randomly sampled trajectories from the R2R-CE benchmark. 
We compare our method with two representative models, StreamVLN \cite{wei2025streamvln} and DualVLN \cite{wei2025ground}, both of which are built upon an 8B-scale backbone.
Specifically, DualVLN suffers from high latency because it necessitates processing 9 distinct images at each inference step. 
Meanwhile, although StreamVLN utilizes a caching mechanism, its overall speed is hindered by the periodic recomputation required to refresh its global memory.
As shown in Table~\ref{tab:latency}, our \model{} achieves the lowest inference latency of 159\,ms, significantly outperforming both DualVLN (472\,ms) and StreamVLN (221\,ms). 

\myparagraph{Component-wise Ablation.}
\label{sec:exp_component}
Table~\ref{tab:component_ablation} provides the ablation for each component, starting from a vanilla baseline (constant DAgger mixing ratio, single-step supervision, and KV-cache reuse at inference).
1) Replacing the constant DAgger schedule with our Ada. DAgger schedule lifts R2R-CE SR by $+7.5$ points (60.3$\to$67.8) and RxR-CE SR by $+11.8$ points (63.8$\to$75.6).
This confirms the motivation that early exploration in the rollout reshapes the training distribution toward recoverable states far more effectively than a uniform schedule.
2) Utilizing packed training further advances R2R-CE SR to 73.2 and SPL to 65.0, while maintaining RxR-CE SR at 75.6.
3) KV-Cache Reuse delivers a $5.6\times$ latency reduction, decreasing from 898\,ms to 159\,ms.
These efficiency gains occur with negligible performance fluctuations.

\begin{figure*}[t]
\centering
\includegraphics[width=1.0\textwidth]{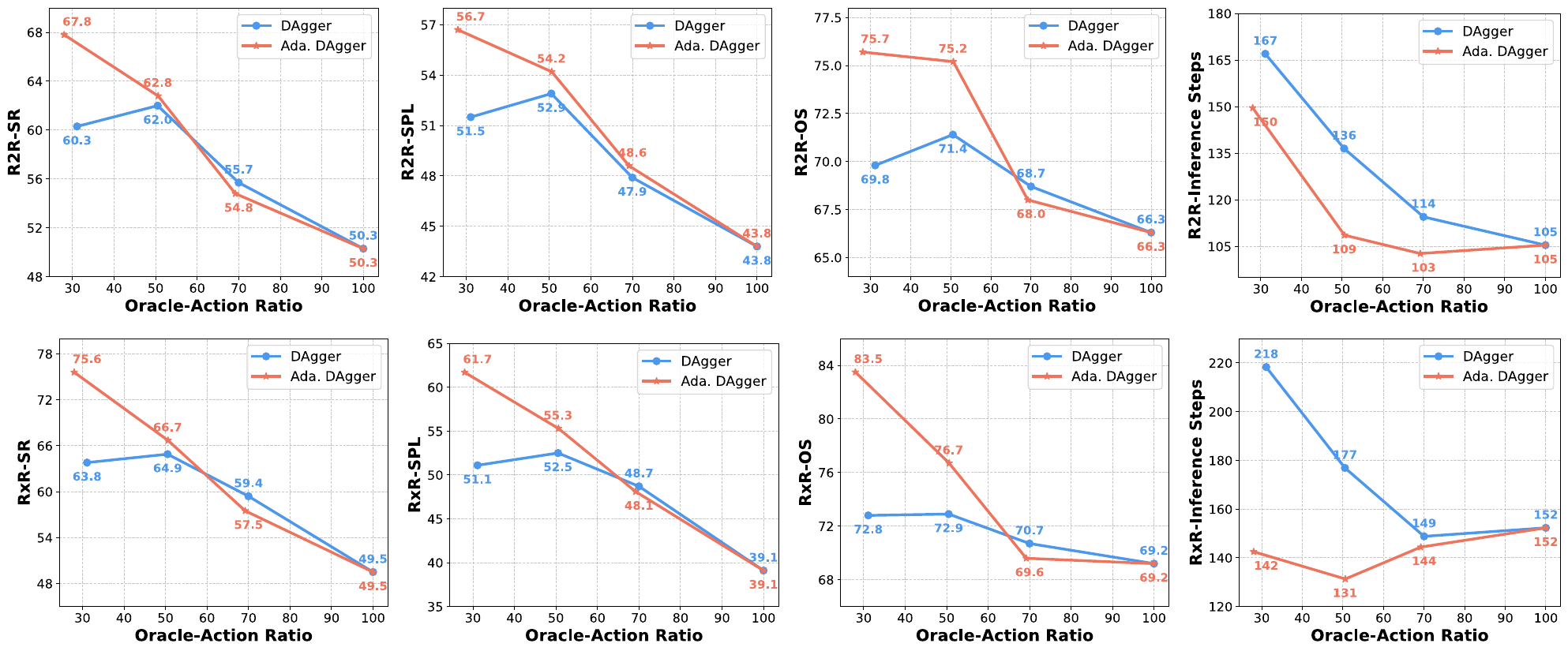}
\caption{
\textbf{Comparison of Vanilla DAgger (constant mixing ratio $\beta\in\{0.3,0.5,0.7\}$) and our Adaptive DAgger ($\alpha\in\{0.02,0.2,0.6\}$) on R2R-CE (top) and RxR-CE (bottom).} Each method is plotted against the \emph{Oracle-Action Ratio} (the average ratio of oracle-guided actions per rollout) so that the two schedules are compared at matched exploration level. The right-most point ($100\%$) corresponds to plain imitation learning.
}
\label{fig:dagger_comp}
\end{figure*}

\paragraph{Effect of Sampling Strategy.}
Table~\ref{tab:sampling} evaluates all combinations of the two frame-sampling strategies across training and test phases. (1)~Models trained with uniform sampling consistently outperform those trained with the sliding window strategy, which we attribute to its more balanced spatiotemporal coverage of the trajectory history. For instance, when evaluated with uniform sampling, the uniform-trained model achieves significant absolute SR improvements of 9.4 ($63.8 \to 73.2$) on R2R-CE and 7.4 ($68.2 \to 75.6$) on RxR-CE compared to its sliding-window-trained counterpart. (2)~Within the uniform-training rows, switching from uniform to sliding-window sampling at test time incurs only a marginal drop in SR ($73.2 \to 72.2$ on R2R-CE, $75.6 \to 74.1$ on RxR-CE), confirming that our model is not tied to the training-time sampling policy. This test-time flexibility enables the choice of the sampling strategy that best fits task settings without any retraining cost.

\myparagraph{Comparison of Training Schemes.}
Table~\ref{tab:attention_mask} compares three training schemes, including ``single-step'', ``naive packing'' (\ie, packed + causal mask), and our proposed ``packed + action-isolating mask''.
To ensure a fair comparison under a strictly matched optimization budget, all three variants are trained for exactly 30K iterations.
Comparing Rows 1 and 2, we observe a drop in SR, decreasing from 67.8 to 62.3 on R2R-CE, and from 75.6 to 61.8 on RxR-CE. 
This confirms that naively packing the data heavily degrades navigation performance due to severe action leakage between sequential actions.
By employing the action-isolating mask to explicitly prevent this leakage, we observe remarkable performance gains. 
Specifically, compared to naive packing, our proposed method yields a substantial absolute improvement of +10.9 (from 62.3 to 73.2) in SR on R2R-CE. 
On RxR-CE, it delivers an even more renamrkable +13.8 surge (from 61.8 to 75.6), effectively matching the upper bound of the ``single-step'' baseline. 
These results clearly demonstrate the effectiveness of our proposed packed training schema.

\myparagraph{Comparison between Adaptive and Vanilla DAgger.}
\label{sec:exp_dagger}
We compare Vanilla DAgger (constant mixing ratio $\beta \in \{0.3, 0.5, 0.7\}$) against our Adaptive DAgger (decay coefficient $\alpha \in \{0.02, 0.2, 0.6\}$). To ensure a fair comparison, we align the two grids based on their \emph{empirical expert ratios}, the average fraction of oracle-guided actions per rollout, at roughly 30\%, 50\%, and 70\%. 

When the expert ratio is high (~70\%), both methods are heavily guided by the oracle, behaving similarly to plain imitation learning, which limits the collection of diverse self-correction states. However, as the expert ratio drops to ~30\%, forcing the agent to explore its own policy more extensively, the superiority of our adaptive schedule becomes highly pronounced. The performance of Adaptive DAgger climbs monotonically, reaching 67.8 SR on R2R-CE and 75.6 SR on RxR-CE (an absolute improvement of +7.5 and +11.8 over the best Vanilla DAgger point at matched exploration levels). 
At a 30\% expert ratio, Vanilla DAgger's inference steps escalate prohibitively (167 on R2R-CE, 218 on RxR-CE) while its SR ceases to improve. Conversely, our adaptive schedule effectively bounds the inference steps (150 on R2R-CE, 142 on RxR-CE) by permitting early exploration while ensuring oracle guidance in later steps. This bounded trajectory not only secures SR and SPL but also translates directly into fewer inference steps, effectively aligning with our core objective of efficient navigation.

\vspace{-2.5mm}
\section{Conclusion}
\label{sec:conclusion}
In this paper, we present \model{}, a highly efficient and robust baseline that fundamentally addresses the systemic bottlenecks constraining current MLLM-based VLN. By systematically dissecting the prevailing inefficiencies across inference, training, and data collection, we introduce a coherent recipe consisting of three simple-yet-effective mechanisms. Specifically, our \textit{KV-cache reuse with contiguous RoPE} fundamentally eliminates computational redundancy, enabling strictly constant latency for real-time streaming. Meanwhile, the \textit{packed training with an action-isolating mask} effectively bridge the training-inference gap by preventing actinon leakage, and the \textit{Adaptive DAgger} optimally balances autonomous exploration with sample efficiency. 
Experimental results on the R2R-CE and RxR-CE benchmarks show that \model{} achieves state-of-the-art performance while reducing per-step inference latency by 82\% (from 898\,ms to 159\,ms) compared to the same model without cache. We hope the systematic design principles established in \model{} can serve as a robust foundation for future streaming VLN agents.


\bibliography{egbib}
\bibliographystyle{plain}


\appendix

\section{Limitations}
\label{sec:limitations}
While \model{} significantly advances the efficiency and performance of streaming VLN agents, several limitations remain to be addressed in future research. 

First, our current evaluations are conducted entirely within simulated environments (\ie, R2R-CE and RxR-CE). Despite the high visual fidelity of these simulators, real-world deployment entails unpredictable physical dynamics—such as motion blur, dynamic obstacles, and complex lighting variations—which may challenge the robustness of our current visual representations. Bridging this sim-to-real gap remains a critical next step. 

Second, to achieve ultra-low inference latency, \model{} deliberately relies strictly on egocentric monocular RGB inputs. While this design is highly efficient and scalable, the absence of a panoramic field-of-view (FOV) or explicit depth geometry (\eg, RGB-D) may occasionally lead to suboptimal navigation decisions in visually ambiguous or expansive open-space scenarios, where broader spatial awareness is highly beneficial.

Finally, although our continuous KV-cache reuse mechanism and frame sampler effectively eliminate computational redundancy and maintain constant latency, the memory footprint (VRAM) of the KV-cache still grows linearly with the number of retained historical states. For extremely long-horizon tasks spanning thousands of steps, this could eventually pose memory constraints on edge robotic devices. Future work will explore more aggressive cache eviction or token compression strategies to further accommodate infinite-horizon navigation.

\section{Broader Impact and Code of Ethics}
\label{sec:broader_impact}

Our research advances the efficiency and real-time responsiveness of Vision-Language Navigation, a core capability for embodied AI. The ability for robots to rapidly and accurately interpret natural language instructions and visual surroundings has profound positive implications. It can significantly accelerate the development of domestic service robots, search-and-rescue agents in hazardous environments, and navigational aids for the visually impaired.

However, the deployment of such vision-language agents raises important ethical and safety concerns. \model{} continuously processes egocentric RGB streams of indoor, often private, environments (\eg, homes and offices). If deployed on mobile agents, this presents significant privacy risks regarding unauthorized data collection and surveillance. To mitigate this, future real-world systems must prioritize local, on-device processing to prevent sensitive domestic video data from being transmitted to cloud servers. Furthermore, because our agent is built upon a pre-trained Multimodal Large Language Model, it may inherit the inherent biases and hallucinations of its foundational backbone. In physical deployments, an agent misinterpreting a spatial command or failing to recognize a safety-critical obstacle (\eg, stairs or fragile objects) could result in property damage or physical harm. Consequently, implementing robust safety guardrails and uncertainty-aware fallback mechanisms is essential before transferring this technology to physical, human-centric environments.





\end{document}